\documentclass[11pt]{article}
\usepackage{spconf,amsmath,amssymb,graphicx}
\usepackage{url}
\usepackage{float}
\usepackage{caption}
\usepackage[ruled,vlined]{algorithm2e}
\usepackage{booktabs}
\usepackage[colorlinks=true,linkcolor=blue,citecolor=blue,urlcolor=blue]{hyperref}
\usepackage{orcidlink}
\usepackage[T1]{fontenc} % Essential for correct character encoding
\usepackage{newtxtext}   % Sets text font to Times
\usepackage{comment} % Enables the comment environment
\usepackage{newtxmath}
\usepackage[colorlinks=true, linkcolor=blue, citecolor=blue, urlcolor=blue]{hyperref}
\title{STOP DENOISING YOUR BLURS}

\name{Sasidhar Parvathireddy \orcidlink{0009-0001-1559-349X},
Vamsidhar Saraswathula \orcidlink{0000-0003-3173-8238},
Rama Krishna Gorthi \orcidlink{0000-0001-5021-0071}
}

\address{Indian Institute of Technology Tirupati, India.}

\begin{document}

\maketitle
\begingroup
\renewcommand\thefootnote{}
\footnotetext{© 2026 IEEE. Personal use of this material is permitted. Permission from IEEE must be obtained for all other uses, in any current or future media, including reprinting/republishing this material for advertising or promotional purposes, creating new collective works, for resale or redistribution to servers or lists, or reuse of any copyrighted component of this work in other works.}
\endgroup    
\begin{abstract}
 In recent times, diffusion models have achieved remarkable performance in image restoration tasks. Their core mechanism relies on the restricted presumption of degradation prior to the additive noise operation. However, the blur model, one of the most widely studied degradation formulations, violates this assumption, as it is inherently based on convolution rather than addition. In this paper, we introduce \textit{ConvDiff}, a novel diffusion based framework that substitutes the additive operation with convolution for the task of image deblurring. In the forward process, we construct a meaningful trajectory from the clean image to its blurred counterpart by exploiting the frequency domain characteristics of convolution, rather than progressively corrupting the image with additive noise. While the current work instantiates this framework for Gaussian blur, where frequency-domain decomposition yields closed-form and physically valid intermediate states, the underlying principle of constructing degradation trajectories from the blur operator extends naturally to other blur families. This formulation bridges the gap between the mathematical principles of blurring and the iterative design of diffusion-based restoration algorithms, enabling more physically grounded and effective image restoration models.

\end{abstract}


\begin{figure*}
    \centering
    \includegraphics[width=0.70\linewidth]{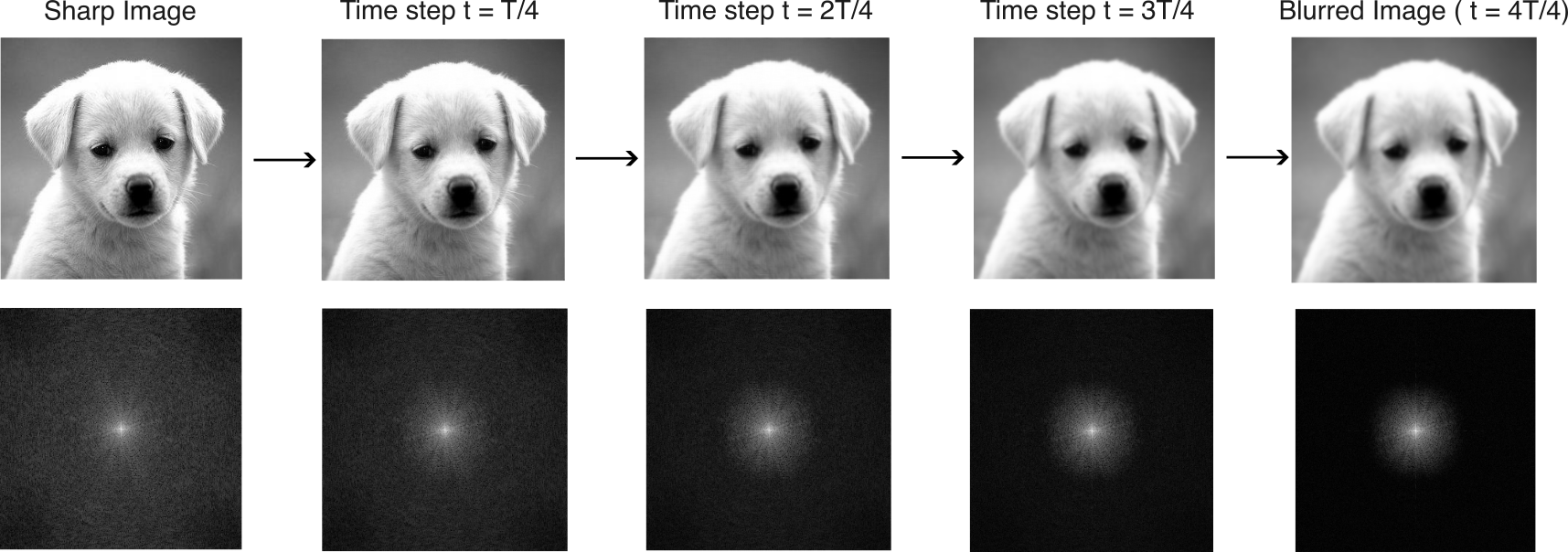}
    \caption{Generated intermediate images (for T= 4) and their corresponding Fourier Transforms between a sharp and blurred image pair (for gaussian blur)}
    \label{fig:motion}
\end{figure*}

\section{Introduction}
Diffusion models \cite{ddpm} have made a remarkable impact in the field of computer vision, achieving unprecedented performance in image generation. The principle behind diffusion models is to generate images by decomposing the complex generation task into a sequence of small, learnable denoising steps. During training, the images are progressively corrupted with Gaussian noise to produce intermediate noisy representations, and the model learns to invert this gradual corruption process to reconstruct the original data distribution.

\par Diffusion models have also been extended to image restoration tasks such as deblurring and super-resolution, where the model is conditioned on a degraded image and learns to generate its clean counterpart from pure noise through iterative denoising, as first demonstrated in the SR3 model \cite{sr3}. However, this approach introduces a fundamental mismatch when applied to image deblurring, where degradation arises from convolution, not additive noise. While diffusion models simulate degradation through additive white Gaussian noise, deblurring involves a blur kernel acting as a convolution operator. Consequently, the intermediate images generated along the diffusion path lack physical significance when viewed from the true degradation perspective.

 \par Works like Cold Diffusion \cite{colddiff} , Inverse Heat Dissipation \cite{Heatdissipation} and Progressive Blur \cite{progressiveblur} substitute Gaussian noise with Gaussian blur producing visually coherent but physically unconstrained intermediate states. However they focus on unconditional image generation rather than true restoration.  
\par Recently, works such as FideDiff \cite{FideDiff} and BlurDM \cite{BlurDM} integrate blur physics into diffusion models by using simulated exposure time to control the strength of blur in each forward step. While this successfully creates meaningful intermediate states, its reliance on an exposure-time restricts its applicability to motion blur. This motivates a more general formulation - one that derives the degradation trajectory directly from the observed blur kernel itself, rather than from any auxiliary physical model.

\par A key challenge in designing a physically consistent iterative deblurring framework is that convolution is inherently a single-step operation, making it difficult to integrate into a multi-step iterative model. Some prior works, such as \cite{fang2014separable}, attempt to decompose a blur kernel into interpretable physical components (e.g., trajectory, intensity, or point spread function). While such approaches offer valuable insight, they still do not yield partially blurred intermediate images, leaving the need for a mechanism that meaningfully bridges sharp and blurred states.

In this work, propose \textit{ConvDiff},  a physically-aware iterative deblurring framework built directly on convolutional operations, which we instantiate and validate for Gaussian blur. Unlike conventional diffusion models that rely on additive noise, \textit{ConvDiff} replaces the noise process with a convolutional mechanism, decomposing a single-step blur into a progressive, multi-step sequence governed by frequency-domain properties of convolution. This formulation enables the generation of physically meaningful intermediate images that bridge sharp and blurred states. We further demonstrate that this generic and physically grounded approach achieves improved restoration performance compared to traditional diffusion-based methods, highlighting the importance of degradation-specific model design.

\section{Proposed Methodology}
%\label{sec:guidelines}
\subsection{Problem formulation for progressive blurring}

Assume $k_{\text{blur}}$ denotes the degradation blur kernel that relates the sharp ($x$) and blurred ($y$) images. The relationship of $x$,$y$, and $k_{\text{blur}}$ in the spatial domain is given in Eq. \ref{eq1}.
\begin{equation}
\label{eq1}
 y = x * k_{\text{blur} }
\end{equation}
Taking Fourier transform of Eq.\ref{eq1} gives,
\begin{equation}
\label{eq2}
Y = X \cdot H_{\text{blur}}
\end{equation}
 Let $\mathcal{F}\{x\}=X ,\ \mathcal{F}\{y\}= Y ,\ \mathcal{F}\{k_{\text{blur}}\}= H_{\text{blur}}$  where $\mathcal{F}\{\cdot\}$ denote the Fourier Transform.
We aim to decompose the overall blur kernel into a sequence of smaller, incremental kernels that together reproduce the same final blur. Owing to the \textbf{\textit{associative property}} of convolution \cite{oppenheim1996signals}, a system can be represented as a cascade of n sequential subsystems whose combined effect is equivalent to that of the original blur kernel.
 \begin{equation}
 \label{eq_blur_expansion}k_{\text{blur}} = k_1 * k_2 * k_3 * ....... * k_n\end{equation}
\par Using the above factorization, we can rewrite the overall blur operation as a sequence of convolutions applied to the sharp image,
\begin{equation}
\begin{split}
    y &= (((x_0 * k_1 ) * k_2 ) * k_3 ) ...... ) * k_n   
\end{split}
\end{equation}
%\subsection{Convolution in 'n' steps}
Let us refer the sharp image $x$ with $x_0$ from here and $x_t$ denotes the intermediate image obtained after $t$ stages of blur. Then we can write :
\begin{equation}
x_t = x_0 * \bar{k_t}, \quad \text{where } \bar{k_t} = k_1 * k_2 * \dots * k_t
\label{kbar}
\end{equation}

Note that $y = x_0 * \bar{k_n}$ and with $\bar{k_t}$ we can calculate any corresponding intermediate image $x_t$. Taking the Fourier transform of Eq. \ref{eq_blur_expansion} gives:

 \begin{equation}
 \label{eq_blur_expansion_freq}
 H_{\text{blur}} = H_1 \cdot H_2 \cdot H_3 \cdot .......\cdot H_n
 \end{equation}
Now, we generate $n$ intermediate images between $x_0$ and $y$, assigning each step an equal fraction of the total blur $H_k$, which is given in Eq.\ref{eq_HK}. 
\begin{equation}
\label{eq_HK}
H_k = H_{\text{blur}}^{\frac{1}{n}}, \quad \forall k \in \{1, 2, \dots, n\}.
\end{equation}

From Eq.\ref{kbar} and Eq.\ref{eq_HK} we can write 
\begin{equation}
    \mathcal{F}\{ \bar{k_t} \} =H_{blur}^{\frac{t}{n}} 
\end{equation}

\begin{equation}
    X_t = X_0 \cdot H_{blur}^{\frac{t}{n}}
    \label{tbyn}
\end{equation}

where $X_t = \mathcal{F}\{{x_t} \}$ and $X_0 =\mathcal{F}\{{x_0} \}$. Let us define $\beta = \frac{t}{n}$, where $\beta$ is degradation strength and is $ 0 < \beta \leq 1 .$ Here $\beta = \frac{t}{n}$ controls the progression of blur. As $\beta \rightarrow 0$, $x_t$ approaches the sharp image $x_0$; as $\beta \rightarrow 1$, it converges to the fully blurred image $y$. The frequency components of the blur kernel are modulated such that their magnitudes are exponentially scaled and their phase terms are linearly scaled by a factor of $\beta$. This controlled attenuation yields a partially blurred image in the spatial domain.

\begin{equation}
H_{\text{blur}}^\beta = |H_{\text{blur}}|^\beta e^{j\phi \beta}
\end{equation}

\par For Eq.\ref{tbyn} to yield physically meaningful intermediate images,\, $H_{blur}^{\beta}$ must remain a valid blur kernel, that is, its inverse Fourier transform should be real, non-negative, normalized, and spatially localized. This is guaranteed for Gaussian blur kernels, because the Fourier transform of a Gaussian is itself Gaussian, and the fractional powers of the Gaussians remain Gaussian.

\par Figure \ref{fig:motion} illustrates the intermediate results obtained for the sharp and blurred image pair using the equations mentioned above. The blur strength is seen to increase progressively from left to right, while the corresponding Fourier spectra demonstrate a systematic reduction in high-frequency content.

\subsection{Progressive deblurring}

 Let us denote the degradation function that generates intermediate images with $D( x_{0},k_{\text{blur}},\beta)$, where $ 0 \leq \beta \leq 1 $.
 
\begin{equation}
   x_\beta =  D( x_{0},k_{\text{blur}} ,\beta) = \mathcal{F}^{-1} \{ X_{0}.H_{\text{blur}}^{ \beta } \} 
\end{equation}
where $\mathcal{F}^{-1}\{\cdot\}$ denotes Inverse Fourier transform. If the blur kernels are available, they can be directly utilized; otherwise, they can be estimated using the Wiener inverse filtering formulation \cite{wiener1949extrapolation} \cite{murli1999wiener} as given in Eq. \ref{wiener}. The Wiener-based estimation approach stabilizes frequency-domain division by introducing a regularization term $S$. In noise-free or synthetic datasets, $S$ can be set to a small positive constant to ensure stable inversion without significantly distorting the estimated kernel. For our dataset, we empirically determined the optimal value of $S$ by evaluating reconstruction performance over a range of candidate values and selecting the one that achieved the best visual fidelity and quantitative accuracy. Specifically, we employed $S = 10^{-8}$ for all experiments.

\begin{equation}
    H_{blur} =  \frac{Y.X^*}{(|X|^2 + S)}
    \label{wiener}
\end{equation}
\par where \(X^*\) is the complex conjugate of X
\section{Experimental setup}
\subsection{Training}
The objective is to design a deep learning model that approximates the inverse of the degradation function \( D \). 
We denote this inverse approximation as \( I_{\theta}(x_{\beta}, \beta) \approx \hat{x}_0 \), 
where the model predicts the sharp image \( x_0 \) given a degraded image \( x_{\beta} \) and its corresponding degradation strength \( \beta \). 
Similar to diffusion models, during training, a random intermediate state \( x_{\beta} \) is generated, and the model learns to reconstruct the original clean image \( x_0 \) from this partially degraded version. 
This process enables the network to learn the direction along which the restoration trajectory should evolve.  

Unlike conventional diffusion models that rely on discrete timesteps, \textit{ConvDiff} assumes a continuous degradation process, where \( \beta \sim \mathcal{U}(0, 1) \) following the formulation of INDI~\cite{indi}. 
The network is therefore trained to predict the final sharp image \( \hat{x}_0 \) from any intermediate degraded state \( x_{\beta} \) sampled along this continuous degradation path.

\begin{algorithm}
\SetAlgoLined
\KwIn{Input sharp image \(x_0\)}
\For{each training sample}{
    Randomly sample \(\beta \sim \mathcal{U}(0, 1]\)\;
    Generate degraded image: \(x_{\beta} = D(x_0, k_{\text{blur}} ,\beta)\)\;
    Predict: \(\hat{x}_0 = I_{\theta}(x_{\beta}, \beta)\)\;
    Compute loss: \(\mathcal{L} = \| \hat{x}_0 - x_0 \|_2^2\)\;
    Back-propagate and update network parameters ($I_{\theta}$) to minimize \(\mathcal{L}\)\;
}
\caption{Training procedure for ConvDiff}
\label{alg:convdiff}
\end{algorithm}

A variant of U-Net with ConvNext \cite{convnet} blocks, similar to the network used in Cold Diffusion \cite{colddiff}, is employed to approximate the inverse degradation function \( I_{\theta}(x_{\beta}, \beta) \).

\subsection{Inference}

\par During the inference stage of diffusion models, the model performs a denoising process: given a noisy sample \( x_t \), it predicts the noise component added at step \( t \). 
This prediction, combined with the known diffusion schedule parameters, is then used to compute a less noisy sample \( x_{t-1} \). 
The intermediate image \( x_{t-1} \) is subsequently fed back into the model, and the procedure is iterated for \( n \) steps to reconstruct the final clean sample \( x_0 \). 

\textit{ConvDiff} follows a similar iterative mechanism but performs \textit{progressive deblurring} in the frequency domain. 
At each time step \( t \), the model \( I \) predicts a cleaner estimate of the target image \( x_0 \). 
We compute the Fourier transforms of both the current prediction \( \hat{x}_0 \) and the initial blurred input image \( y \), 
and estimate a temporary blur kernel \( \tilde{H}_{\text{blur}} \) corresponding to the current step using the kernel approximation function defined in Eq.~\ref{wiener}. 
This operation can be denoted as follws:

\begin{equation}
W\big( \mathcal{F}\{\hat{x}_0\}, \mathcal{F}\{y\} \big) = \tilde{H}_{\text{blur}},
\end{equation}

where \( W(\cdot) \) represents the Wiener-based kernel estimation function from Eq. \ref{wiener}. 
The estimated kernel \( \tilde{H}_{\text{blur}} \) is then used in Eq.~\ref{tbyn} to generate the next intermediate image for the reverse process, 
which is fed back into the model at step \( t-1 \). 
This iterative refinement continues until until \( t = 0 \), 
yielding the final restored sharp image \( \hat{x}_0 \).

\begin{algorithm}
\SetAlgoLined
\KwIn{Blurred image $y$}
Initialize $x_{t} = y$\;

\For{$t = n, n-1, n-2, \dots, 1$}{
    $\hat{x}_{0} = I_{\theta}(x_t, \frac{t}{n})$\; %\quad , \text{where $\frac{t}{n}=\beta$}\; 
    $\Tilde{H}_{\text{blur}} = W( \mathcal{F}\{\hat{x}\},\mathcal{F}\{y\})$\;
    $x_{t-1} = D(\hat{x}_0, \tilde{H}_{\text{blur}}, \frac{t-1}{n})$\;
}
\KwOut{$\hat{x}_0$}
\caption{Inference algorithm for ConvDiff}
\label{alg:inference}
\end{algorithm}
\begin{figure*}[t]
    \centering
    \includegraphics[width=0.7\linewidth]{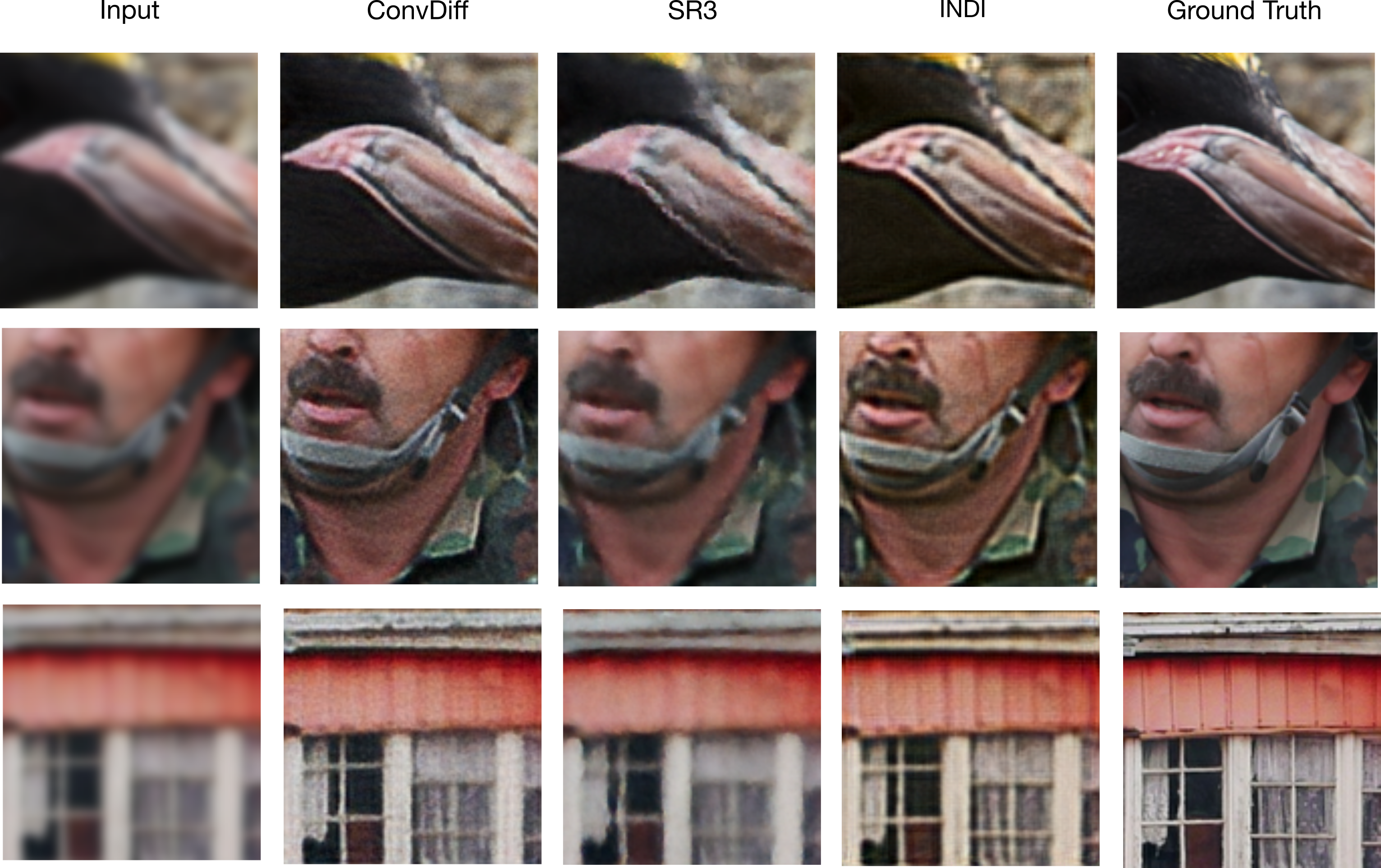}
    \caption{Comparison between ConvDiff and other iterative deblurring frameworks (SR3, INDI).}
    \label{fig:results}
\end{figure*}
\subsection{Dataset details}

For training and evaluation, we used the DIV2K dataset~\cite{agustsson2017ntire}, 
which comprises 800 high-quality natural images at 2K resolution for training 
and 100 images for testing. 
To simulate degradation, each image was convolved with a Gaussian blur kernel 
of size \( 15 \times 15 \), where the standard deviation \( \sigma \) 
was randomly sampled from the range \([2, 4]\). 
From the resulting degraded images, patches of size \( 128 \times 128 \) 
were extracted to construct the training set.

\section{Results}
\noindent
Our primary comparison focuses on two traditional iterative restoration baselines: 
the noise-driven diffusion model SR3~\cite{sr3} and the interpolation-based model INDI~\cite{indi}. 
All three models, \textit{ConvDiff}, SR3, and INDI, were trained using the same U-Net architecture 
augmented with ConvNeXt blocks, ensuring consistency in network configuration and depth across implementations. 
INDI, proposed as an alternative to diffusion models, generates intermediate representations 
via direct linear interpolation between the sharp and blurred images.  

During evaluation, \textit{ConvDiff} was tested with \( n = 5 \) time steps, INDI with \( n = 10 \), 
and SR3 with \( n = 2000 \) steps, following their respective original implementations. 
As illustrated in Fig.~\ref{fig:results}, \textit{ConvDiff} produces sharper and more detailed restorations 
compared to both SR3 and INDI. Quantitative results presented in Table~\ref{tab:results} further support these findings. 
SR3 tends to oversmooth textures and fails to recover fine high-frequency details, 
whereas INDI introduces noticeable block-like artifacts at high magnification. 
In contrast, \textit{ConvDiff} more effectively restores edge sharpness and intricate structures.

\par However, slight residual noise remains in the \textit{ConvDiff} outputs. This is primarily attributed to the iterative Wiener-based kernel estimation used during inference, which may induce minor artifacts that propagate and accumulate through successive time steps. This error propagation likely limits the model's convergence and explains why the quantitative metrics, while superior to traditional diffusion baselines, remain lower than recent end-to-end state-of-the-art (SOTA) methods.

\begin{figure}[t]
    \centering
    \includegraphics[width=0.85\linewidth]{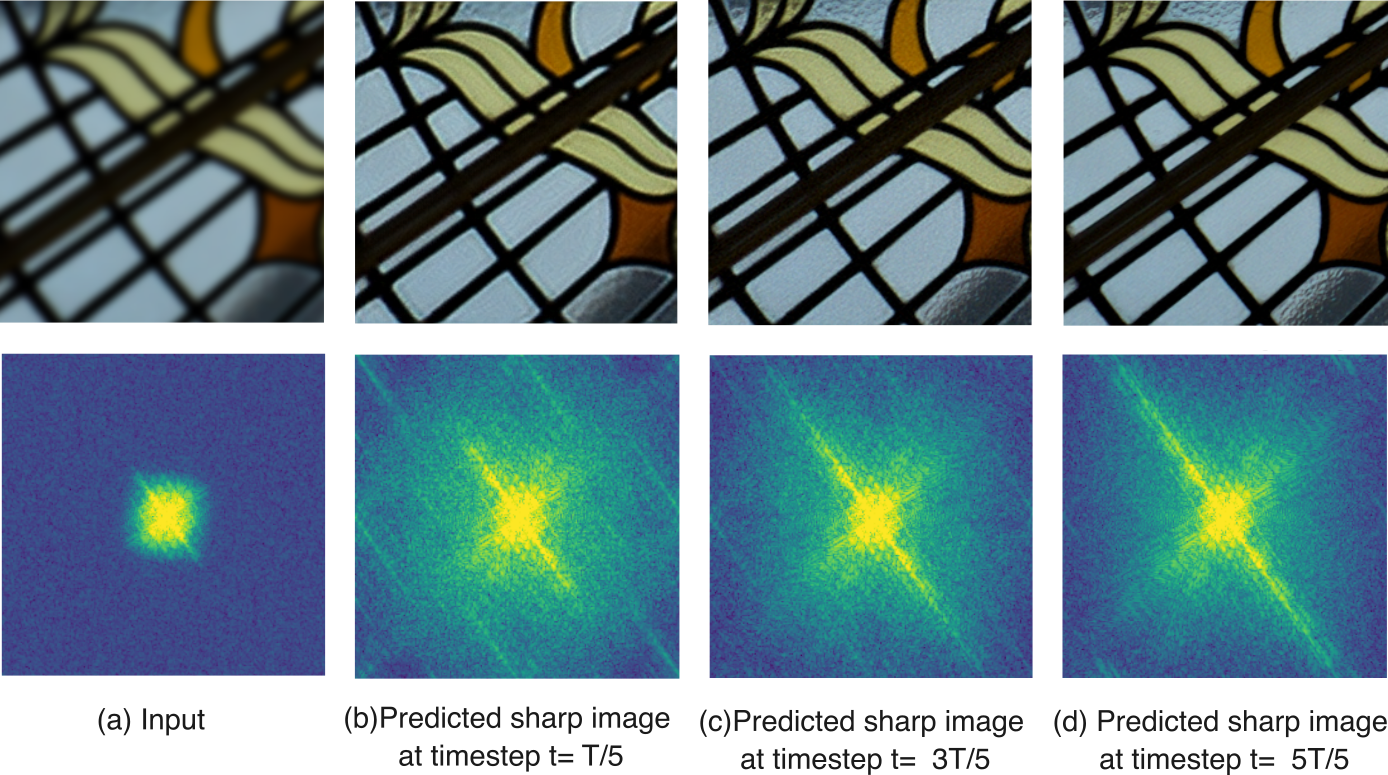}
    \caption{Predicted sharp images and their Fourier Transforms at different time steps.}
    \label{fig:fts}
\end{figure}

We compare \textit{ConvDiff} with several such SOTA restoration networks, including FFTformer~\cite{fftformer}, Restormer~\cite{restormer}, DiffIR~\cite{diffir}, and HiDiff~\cite{hidiff}. These models leverage latent-space priors, multi-scale diffusion mechanisms, and adaptive frequency-domain attention to achieve enhanced performance. Intermediate predictions at successive time steps for our model are shown in Fig.~\ref{fig:fts}, where the gradual recovery of high-frequency components can be clearly seen.

{\captionsetup{labelformat=default}
\begin{table}[h]
    \centering
    \caption{Quantitative comparison: ConvDiff vs. traditional diffusion based methods and other state-of-the-art approaches.}
    \label{tab:results}
    \begin{tabular}{lccc}
        \toprule
        Method & PSNR & SSIM & LPIPS \\
        \midrule
        SR3       & 25.3499 & 0.6747 & 0.3167 \\
        INDI      & 26.0933 & 0.7038 & 0.2073 \\
        \textbf{ConvDiff} & \textbf{29.5927} & \textbf{0.7809} & \textbf{0.1701} \\
        \midrule[\heavyrulewidth]    % first heavy line
        \addlinespace[0.1 pt]           % small gap
        \midrule[\heavyrulewidth]    % second heavy line
        FFTformer & 30.8348 & 0.8760 & 0.1175 \\
        Restormer & 30.9187 & 0.8760 & 0.1337 \\
        DiffIR    & 31.1574 & 0.8795 & 0.1393 \\
        HiDiff    & 31.2655 & 0.8805 & 0.1379 \\
        \bottomrule

    \end{tabular}

\end{table}
}

\section{Conclusion}

To conclude, \textit{ConvDiff} introduces an iterative frequency-domain deblurring framework that progressively reconstructs sharp images by factorizing the blur kernel. This physically interpretable formulation bridges diffusion and convolution based restoration. However, the current formulation has certain limitations. This formulation assumes spatially invariant Gaussian blur, and thus cannot be applied directly on spatially variant degradations like motion blur. Additionally, while inference operates in a blind setting through iterative kernel re-estimation, training requires access to the blur kernels for better estimation of sharp images. The iterative Wiener based re-estimation during inference, while enabling blind restoration, introduces approximation errors that likely account for the residual performance gap relative to end-to-end trained methods. While not yet achieving state-of-the-art performance, \textit{ConvDiff} demonstrates clear improvements in sharpness and structure over conventional diffusion and linear interpolation based methods, highlighting the importance of degradation-specific process design. Future work may explore learned or regularized kernel estimation to replace the current Wiener based approach, extend the framework to spatially variant blur, and adapt the convolutional trajectory to other blur families, for instance, motion blur by parameterizing kernel geometric properties like spread and orientation - pointing toward a broader vision of degradation aware iterative restoration governed by the physics of the degradation itself.

\bibliographystyle{IEEEbib}
\bibliography{strings}

\end{document}